\documentclass[sigconf]{acmart}

\copyrightyear{2026}
\acmYear{2026}
\setcopyright{cc}
\setcctype{by}
\acmConference[MM '26]{Proceedings of the 34th ACM International Conference on Multimedia}{November 10--14, 2026}{Rio de Janeiro, Brazil}
\acmBooktitle{Proceedings of the 34th ACM International Conference on Multimedia (MM '26), November 10--14, 2026, Rio de Janeiro, Brazil}
\acmDOI{10.1145/3767308.3835345}
\acmISBN{979-8-4007-2213-4/2026/11}

\usepackage{amsmath,amsfonts}
\usepackage{enumitem}
\usepackage{graphicx}
\usepackage{xcolor}
\usepackage{booktabs}
\usepackage{multirow}
\usepackage{colortbl}
\usepackage{arydshln}

\begin{document}

\title[Segmentation-Based Attention Entropy for LVLM Hallucinations]
{Segmentation-Based Attention Entropy:
Detecting and Mitigating Object Hallucinations in
Large Vision-Language Models}

\author{Jiale Song}
\affiliation[obeypunctuation]{%
  \department{School of Information and Intelligent Science},
  \institution{Donghua University},
  \city{Shanghai},
  \country{China}
}
\email{2222026@mail.dhu.edu.cn}

\author{Jiaxin Luo}
\affiliation[obeypunctuation]{%
  \department{School of Information and Intelligent Science},
  \institution{Donghua University},
  \city{Shanghai},
  \country{China}
}
\email{1249118@mail.dhu.edu.cn}

\author{Xue-song Tang}
\authornote{Corresponding author.}
\affiliation[obeypunctuation]{%
  \department{School of Information and Intelligent Science},
  \institution{Donghua University},
  \city{Shanghai},
  \country{China}
}
\email{tangxs@dhu.edu.cn}

\author{Kuangrong Hao}
\affiliation[obeypunctuation]{%
  \department{School of Information and Intelligent Science},
  \institution{Donghua University},
  \city{Shanghai},
  \country{China}
}
\email{krhao@dhu.edu.cn}

\author{Mingbo Zhao}
\affiliation[obeypunctuation]{%
  \department{School of Information and Intelligent Science},
  \institution{Donghua University},
  \city{Shanghai},
  \country{China}
}
\email{mzhao4@dhu.edu.cn}

\renewcommand{\shortauthors}{Jiale Song, Jiaxin Luo, Xue-song Tang, Kuangrong Hao, and Mingbo Zhao}

\begin{abstract}
Large Vision-Language Models (LVLMs) achieve strong performance on many multimodal tasks, but object hallucinations severely undermine their reliability. Most existing studies focus on the text modality, attributing hallucinations to overly strong language priors and insufficient visual grounding. In contrast, we observe that abnormal attention patterns within the visual modality can also give rise to hallucinated objects. Building on this observation, we propose \textbf{S}egmentation-based \textbf{A}ttention \textbf{E}ntropy (SAE), which leverages semantic segmentation to quantify visual attention uncertainty in a semantic space. Based on SAE, we further design a reliability score for hallucination detection and an SAE-guided attention adjustment method that modifies visual attention at inference time to mitigate hallucinations. We evaluate our approach on public benchmarks and in real embodied multimodal scenarios with quadruped robots. Experiments show that SAE reduces hallucinations without additional training, improving LVLM reliability.

\end{abstract}



\maketitle

\section{Introduction}
\label{sec:intro}

Large Vision-Language Models (LVLMs) have achieved strong performance on image captioning, visual question answering, multimodal dialogue, and embodied robotic perception~\cite{liu2023visual,li2023blip2,gu2022vision,liu2024improved}. However, they still suffer severely from hallucinations, especially object hallucinations where the model describes objects that do not exist in the image~\cite{rohrbach2018object,li2023pope}.

To mitigate hallucinations in LVLMs, prior work has explored robust visual instruction tuning for better multimodal alignment~\cite{Liu2024LRVInstruction,Jiang2024HACL}, as well as training-free decoding strategies such as Visual Contrastive Decoding (VCD)~\cite{Leng2024VCD}. Other studies have further analyzed the causes of object hallucinations and concluded that language bias is a primary factor~\cite{Huang2024OPERA,Liu2025PAI,liu2024survey}. More recently, some work has begun to investigate visual-side internal mechanisms in LVLMs~\cite{Jiang2025DevilsMiddleLayers}, and has further linked LVLM hallucinations to uncertainty in the model's internal hidden states~\cite{xu2026thinking}. Although these studies have shown effectiveness, they still lack a characterization of the uncertainty of visual attention. In our analysis, we observe that the occurrence of hallucinations is closely related to whether visual attention is concentrated or dispersed, and that more dispersed visual attention often indicates a higher risk of hallucination.

\begin{figure}[t!]
    \centering
    \includegraphics[width=\columnwidth]{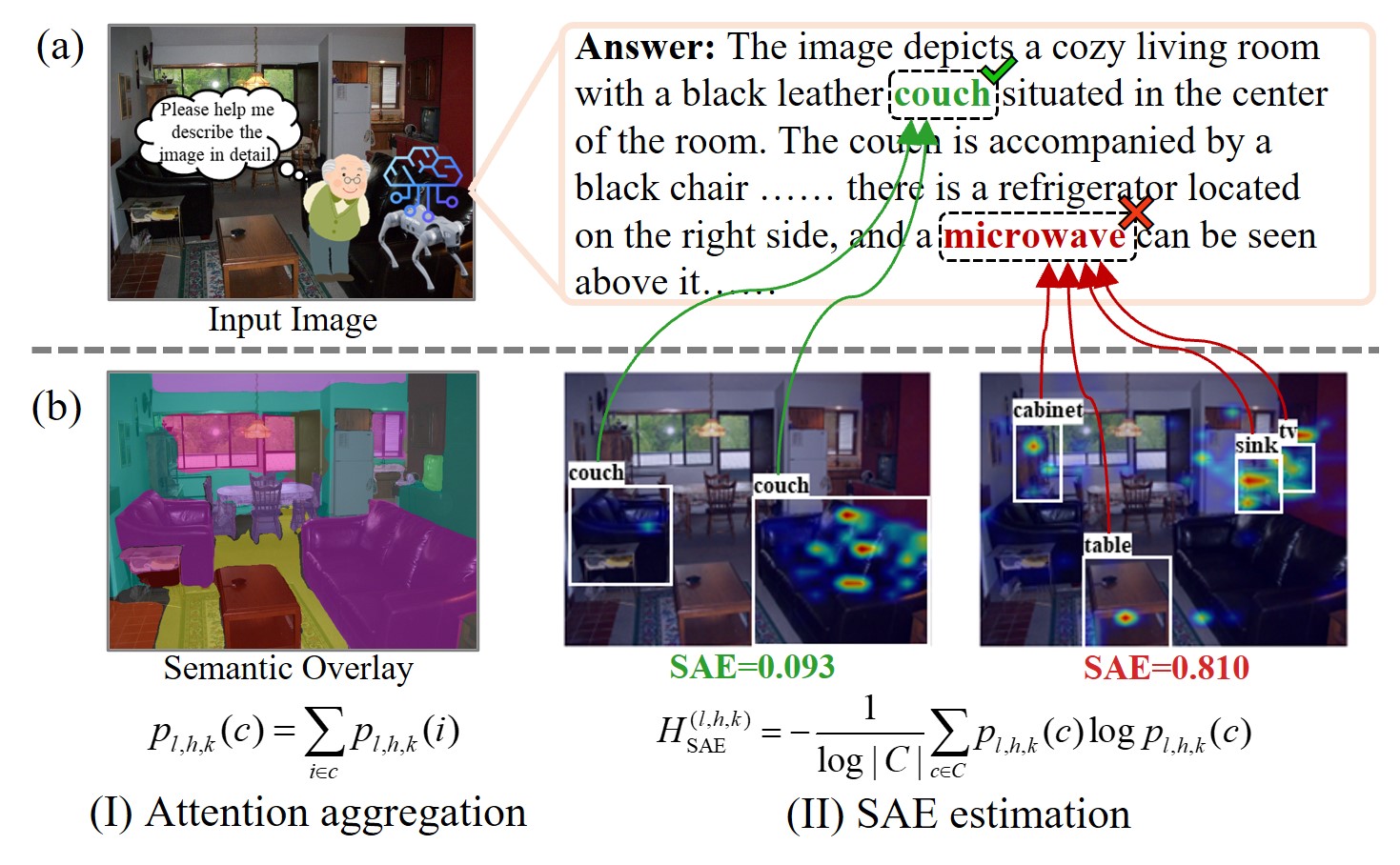}
    \caption{Object hallucinations in LVLMs and SAE. (a) Example of real (green) and hallucinated (red) objects in an LVLM description. (b) SAE pipeline: real objects correspond to low SAE, hallucinated objects to high SAE.}
    \label{fig:intro}
\vspace{-6mm}
\end{figure}

As illustrated in Fig.~\ref{fig:intro}, when an LVLM generates a description, its attention should focus on image regions that are semantically consistent with the target object, rather than being diffused over many irrelevant areas. In practice, regions where visual attention is highly dispersed and which therefore exhibit high attention uncertainty are more likely to give rise to hallucinations.

We propose Segmentation-based Attention Entropy (SAE) to characterize the uncertainty of visual attention inside LVLMs from a semantic perspective. As shown in Fig.~\ref{fig:intro}(b), we apply a semantic segmentation model to the input image, aggregate visual-token attention according to semantic categories, and then compute a normalized entropy over the category-level attention distribution to measure how dispersed the attention is over semantic regions. Empirically, when predicting tokens corresponding to real objects, attention concentrates on semantically consistent regions, leading to low SAE; hallucinated objects, instead, show attention drifting to irrelevant regions and thus significantly higher SAE.

Building on this observation, we construct a reliability score that combines cross-modal attention quality with SAE for object hallucination detection, and design an SAE-guided modulation strategy that directly adjusts attention logits during inference to mitigate hallucinations without any additional training cost. We further validate the proposed method in a quadruped-robot navigation setting, showing its potential for reducing hallucinations in real-world embodied decision-making.

The main contributions of this paper are summarized as follows:
\begin{itemize}[leftmargin=*]

 \item We propose SAE, which quantifies the uncertainty of LVLM visual attention at the object level via semantic segmentation, turning the question of whether attention focuses on semantically consistent regions into a computable entropy metric.
 \item We reveal a correlation between SAE and object hallucinations, and introduce an attention reliability score that fuses cross-modal attention quality with SAE for object hallucination detection.
 \item We develop an SAE-guided inference-time attention modulation method that reshapes visual attention without modifying model parameters, reducing hallucinations across multiple benchmarks and in embodied quadruped-robot scenarios.
 
\end{itemize}

\section{Related Work}
\label{sec:related_work}

\subsection{Large Vision-Language Models}

Large vision-language models (LVLMs) have achieved strong performance on image captioning, visual question answering, multimodal dialogue, and embodied perception by combining pretrained visual encoders with large language models~\cite{chen2024internvl2,li2023blip2,liu2023visual,wang2024qwen2vl}. Recent advances have further improved LVLMs through scaling, alignment tuning, and task adaptation~\cite{liu2024improved,li2024llava,caffagni2024revolution,yan2025task,chen2024dress}. Despite this progress, object hallucination remains a major obstacle to the factuality and reliability of LVLM outputs~\cite{liu2024survey,guan2024hallusionbench,Kaul_2024_CVPR,chen2024multiobject,gong-etal-2024-damro}.

\subsection{LVLM Hallucination and Uncertainty}

Object hallucination has been extensively studied in image captioning and LVLMs, with CHAIR and POPE as evaluation benchmarks~\cite{rohrbach2018object,li2023pope,sun2024mmhal}. Existing mitigation methods include training-time alignment and preference optimization, as well as inference-time decoding strategies such as contrastive decoding, logit recalibration, and visual-attention-based intervention~\cite{zhao2025beyond,Leng2024VCD,Huang2024OPERA,Liu2025PAI,manevich2024lcd,liu2025mfcd,wu2024logical,shang2025pixels,wan2025only}. Recent work has linked hallucinations, uncertainty, and internal model states in LVLMs~\cite{farquhar2024detecting,Li2024ReferenceFreeHallucination,Jiang2025DevilsMiddleLayers,Jiang2025InternalConfidence}. In contrast, visual attention uncertainty remains less explored.

\subsection{Attention Interpretability and Internal Information Flow}

Recent work studies LVLM hallucination through attention patterns, visual grounding, and internal information flow. Analyses of decoder dynamics and attention heads show that hallucinations are closely related to middle-layer visual processing, head-level visual sensitivity, and attention sinks or outlier tokens~\cite{Jiang2025DevilsMiddleLayers,he-etal-2025-cracking,gong-etal-2024-damro}. Other methods mitigate hallucinations by assembling global and local attention, constraining information flow, reweighting image-token attention, or intervening during decoding to strengthen visual grounding~\cite{an2024agla,bai2025adavib,Huang2024OPERA,Liu2025PAI,Leng2024VCD,xu-etal-2025-mitigating,seo2026epistemic}. Building on these attention-based insights, we propose Segmentation-based Attention Entropy, which quantifies visual attention uncertainty at the semantic object level and provides a unified signal for both hallucination detection and inference-time mitigation.

\begin{figure*}[h]

\centering
\includegraphics[width=1\linewidth]{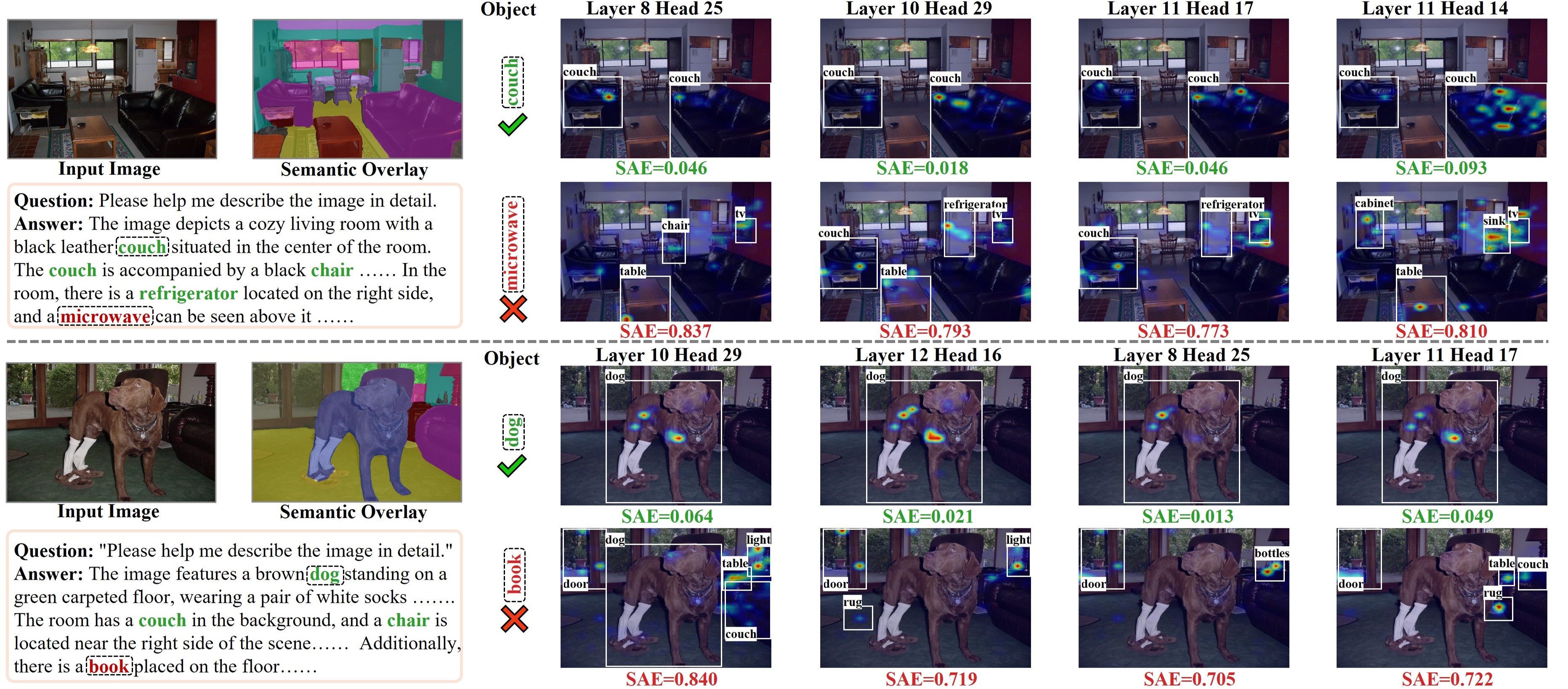}
\vspace{-3mm}
\caption{Head-level attention distributions for real objects (green) and hallucinated objects (red). Real objects have low SAE with attention concentrated on the corresponding image tokens, whereas hallucinated objects have high SAE with attention spread over many irrelevant image tokens.}
\label{fig:vis_layer_head}
\vspace{-3mm}
\end{figure*}

\section{Method}
\label{sec:method}

\subsection{Preliminaries}

\textbf{Notation and attention.}
We discretize the input image into $n$ visual tokens $v_{1:n}$ and the instruction into $m$ text tokens $t_{1:m}$.
At decoding step $k$, the input prefix to the language model is given by
$x_{1:a_k} = (v_{1:n},\, t_{1:m},\, y_{1:k-1})$, where $a_k = n + m + k - 1$.
The model then predicts the next token $y_k$ at this step.
Let $L$ denote the number of Transformer layers and $H$ the number of attention heads.
We denote by $A_k^{(l,h)} \in \mathbb{R}^{a_k \times a_k}$ the attention matrix at layer $l \in [L]$ and head $h \in [H]$ for step $k$, 
where $a_k$ is the prefix length.
In particular, the entry $A_k^{(l,h)}(a_k, i)$ represents the attention weight assigned by the newly generated token $y_k$ to the $i$-th token in the prefix.
When $i$ corresponds to a visual token $v_i$, $A_k^{(l,h)}(a_k, i)$ measures how much attention $y_k$ places on the $i$-th visual token.

\textbf{Background on uncertainty estimation.}
In information theory, Shannon entropy is used to quantify the uncertainty of a discrete distribution. 
Given a discrete random variable with probability vector $p = (p_1,\ldots,p_K)$, its Shannon entropy is defined as
\begin{equation}
H(p) = -\sum_{i=1}^{K} p_i \log p_i.
\label{eq:entropy}
\end{equation}

Existing research typically relates the uncertainty of large language models (LLMs) to hallucinations and employs entropy-based uncertainty measures to detect hallucinated outputs~\cite{farquhar2024detecting}. In the context of LVLMs, however, the allocation of attention in multimodal interactions is particularly crucial~\cite{Jiang2025DevilsMiddleLayers}, yet prior work tends to focus more on the text modality while paying insufficient attention to the image modality~\cite{Huang2024OPERA,Liu2025PAI}. Inspired by these studies, we observe that SAE can be used to quantify the uncertainty of attention distributions within the image modality, which in turn can be exploited to detect and mitigate hallucinations in LVLMs.

\subsection{Segmentation-based Attention Entropy}
\label{sec:method_SAE}

\textbf{Semantic uncertainty in images.}
If we simply treat, for each generated text token, the attention weights over all visual tokens as a probability distribution, we inevitably mix two different sources of uncertainty:
(i) the uncertainty over the underlying visual semantics, and
(ii) the uncertainty induced by the number and size of the tokenized instances that express those semantics.
For example, when a high-confidence object is present in the image, it may correspond either to a single large region covered by many visual tokens or to multiple smaller instances.
In this case, a spread-out attention distribution over visual tokens does not necessarily indicate semantic ambiguity, but may instead reflect the spatial extent or multiplicity of the object.

Our method aims to estimate LVLMs' attention uncertainty over visual semantics when generating a text-level object token, rather than over the number or size of its visual instances. Therefore, we first cluster visual tokens according to semantic segments and then compute attention uncertainty over these clusters. Specifically, the procedure consists of the following two steps:

\textbf{(I) Attention aggregation}

For the $k$-th generated token $y_k$, let $A_k^{(l,h)}(a_k,i)$ denote the attention weight on visual token $v_i$ at layer $l$ and head $h$. We normalize these weights over visual tokens to obtain a probability distribution:
\begin{equation}
p_k^{(l,h)}(i)
=
\frac{A_k^{(l,h)}(a_k,i)}
{\sum_{j=1}^{n} A_k^{(l,h)}(a_k,j)},
\quad i \in \{1,\ldots,n\}.
\label{eq:token-attn}
\end{equation}
We use Mask2Former with a Swin-L backbone \cite{cheng2022masked} for semantic segmentation, obtaining $|C|$ disjoint semantic classes. Each visual token is assigned the semantic label occupying the largest number of pixels within its receptive region. For a semantic class $c \in C$, we aggregate the probabilities of all visual tokens belonging to that class to obtain a class-level distribution:
\begin{equation}
p_k^{(l,h)}(c)
=
\sum_{i \in c} p_k^{(l,h)}(i).
\label{eq:class-attn}
\end{equation}
In other words, the attention probabilities of all tokens corresponding to the same object category (possibly spanning multiple instances and sizes) are pooled into a single semantic class. This aggregation removes the spurious entropy increase caused by spatial extent or repeated same-class regions, making the uncertainty measure more reflective of semantic ambiguity.

\textbf{(II) SAE estimation}

Given the class-level attention distribution in \eqref{eq:class-attn}, we define the SAE as:
\begin{equation}
\mathrm{SAE}_k^{(l,h)}
=
-\frac{1}{\log |C|}
\sum_{c \in C}
p_k^{(l,h)}(c)\log p_k^{(l,h)}(c).
\label{eq:sae}
\end{equation}
This quantity measures how dispersed the attention is across different semantic segmentation classes when predicting $y_k$.
To make SAE comparable across images with different numbers of semantic classes, we normalize by the maximum entropy $\log |C|$, so that $\mathrm{SAE}_k^{(l,h)} \in [0,1]$.

\subsection{Object Uncertainty Estimation with SAE}
\label{sec:SAE_for_Quantifying}

\textbf{Empirical observations and intuition.}
We are interested in whether real and hallucinated objects exhibit systematic differences in their attention distributions within the visual modality.
According to the SAE defined in Sec.~\ref{sec:method_SAE}, when the attention used to predict an object token is more dispersed across semantic segmentation classes, the corresponding SAE is higher, indicating that the visual evidence is less concentrated and the semantic uncertainty is higher, which in turn suggests a higher risk of hallucination.

Fig.~\ref{fig:vis_layer_head} visualizes attention heatmaps and the associated SAE values for several attention heads. When attention mainly focuses on the sofa or dog regions, the SAE values are low; when attention drifts across multiple unrelated regions, SAE becomes much larger. This aligns with our intuition: the more dispersed the attention, the higher the SAE and the greater the hallucination risk.

\begin{figure}[h]
\centering
\includegraphics[width=1\linewidth]{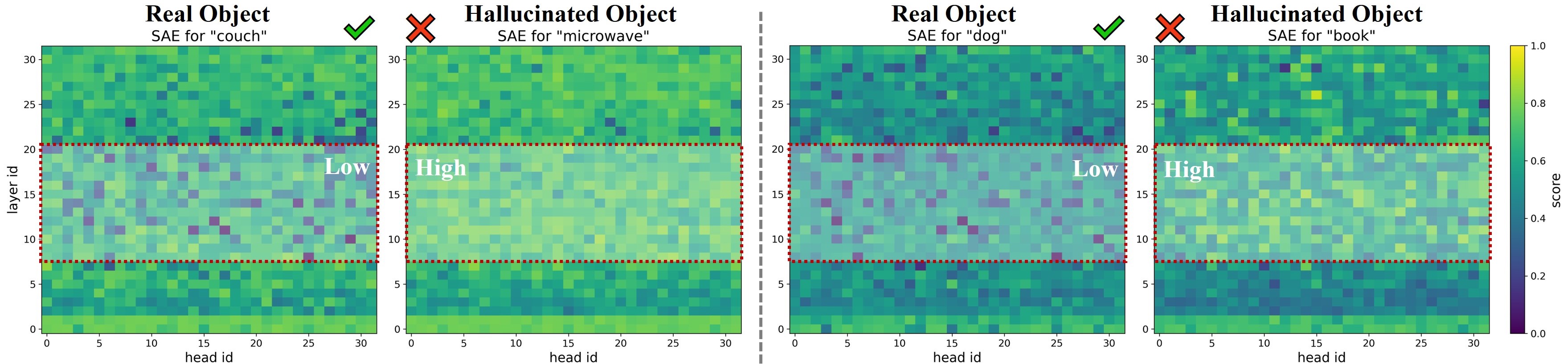}
\caption{Layer-head SAE statistics. The horizontal axis denotes the head index and the vertical axis the layer index, and the difference between real and hallucinated objects is most pronounced in the middle layers.}
\label{fig:vis_low_high}
\end{figure}

Fig.~\ref{fig:vis_low_high} shows SAE values aggregated over all layers and heads of LLaVA-1.5-7B~\cite{liu2024improved} for the real and hallucinated objects in the two examples of Fig.~\ref{fig:vis_layer_head}. A consistent pattern emerges across layer-head pairs: real objects tend to have lower SAE, while hallucinated objects exhibit higher SAE. The gap is most pronounced in the intermediate layers: early layers mainly capture low-level and coarse-grained representations, while very late layers are dominated by language modeling, so the difference diminishes again.

\textbf{Summary and usage.}
Based on the above observations, we treat SAE as a signal that is complementary to cross-modal attention quality for hallucination detection.
Specifically, together with SAE we compute, for the same layer-head, the total attention mass from the text-level object token to all visual tokens, and define an attention reliability score as:
\begin{equation}
R_k^{(l,h)} = M_k^{(l,h)} \bigl(1 - \mathrm{SAE}_k^{(l,h)}\bigr),
\label{eq:reliability}
\end{equation}
where $M_k^{(l,h)}$ is the attention mass, that is the sum of attention weights from the object token to all visual tokens, and $R_k^{(l,h)}$ captures the reliability of both cross-modal and intra-modal attention.
We use a multiplicative form so that reliable evidence requires both sufficient cross-modal attention mass and low semantic dispersion; if either factor is weak, the final reliability remains low.
Intuitively, a high $R_k^{(l,h)}$ indicates that the model is looking at the image in a focused and confident manner.
Following the empirical pattern observed in Fig.~\ref{fig:vis_low_high} and the middle-layer design in VAR~\cite{Jiang2025DevilsMiddleLayers}, we aggregate the reliability scores by averaging $R_k^{(l,h)}$ over middle layers for each head, and then averaging the resulting values over all attention heads to obtain a scalar score for hallucination detection.

\subsection{SAE-guided Object Hallucination Mitigation}
\label{sec:sae_guided_mitigation}

Sec.~\ref{sec:SAE_for_Quantifying} shows that the more dispersed the attention used to predict an object token is across semantic segmentation classes, the higher the corresponding SAE, and the more likely the token is to be hallucinated.
To reduce the hallucination rate of the model, we therefore aim to adjust the attention distribution over visual tokens so that the model focuses more on cross-head consistent, image-relevant regions rather than spreading attention across unrelated areas.

\begin{figure*}[!t]
\centering
\includegraphics[width=1.0\linewidth]{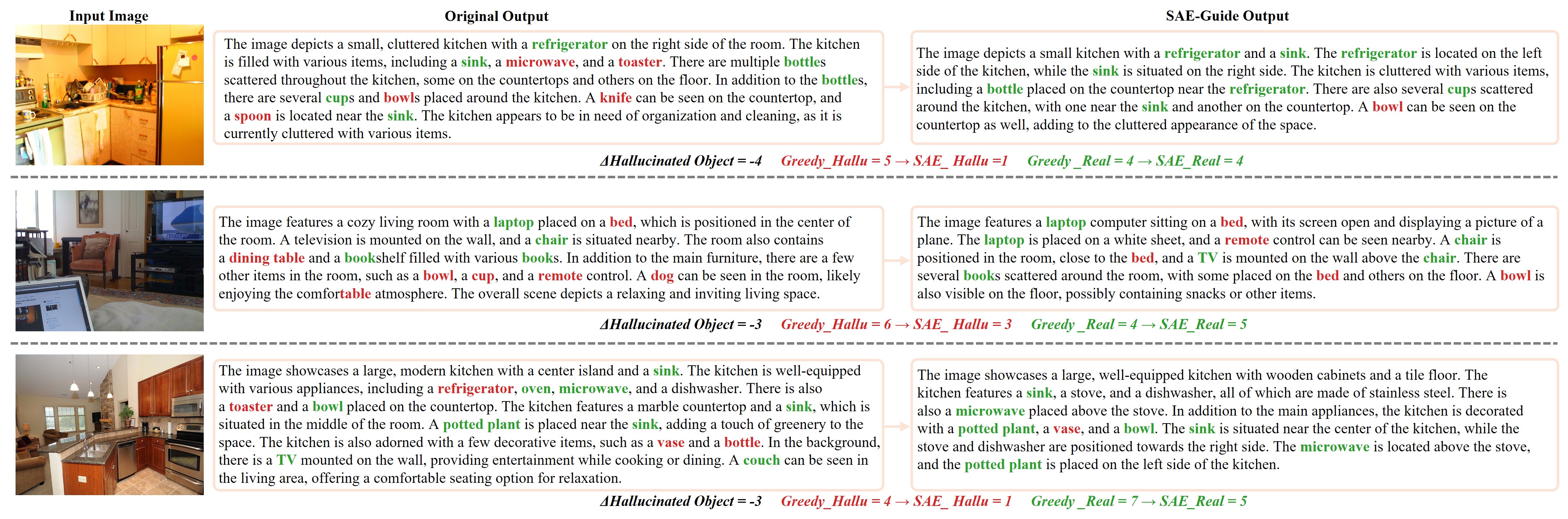}
\caption{Qualitative example of SAE-guided attention intervention during inference. SAE-guided modulation reduces hallucinated objects (red) such as a spoon, a dining table, and a dog, while preserving or even completing real objects (green).}
\label{fig:sae_mitigation}
\vspace{-3mm}
\end{figure*}

We design an SAE-guided attention modulation strategy that intervenes in the attention scores (pre-softmax logits) during inference.
Let $S_k^{(l,h)}(a_k,i)$ denote the pre-softmax attention logit at decoding step $k$ from the query position $a_k$ to the $i$-th visual token at layer $l$ and head $h$, where $i \in \{1,\ldots,n\}$.
We then construct an attention consistency map by taking the absolute values of these logits and averaging them over heads:
\begin{equation}
C_k^{(l)}(i)
=
\frac{1}{H} \sum_{h=1}^{H} \bigl| S_k^{(l,h)}(a_k,i) \bigr|,
\quad i \in \{1,\ldots,n\}.
\label{eq:consistency}
\end{equation}
We use the absolute pre-softmax logits as a sign-agnostic proxy for cross-head interaction strength, avoiding cancellation between positive and negative logits. Heads with high SAE are considered less semantically certain, and are therefore more strongly regularized toward this shared steering pattern. For each head, we then compute its SAE at step $k$ and use it as a linear gate to determine a head-specific enhancement strength:
\begin{equation}
\tilde{S}_k^{(l,h)}(a_k,i)
=
S_k^{(l,h)}(a_k,i)
+
\lambda\, \mathrm{SAE}_k^{(l,h)}\, C_k^{(l)}(i),
\label{eq:sae_update}
\end{equation}
where $\lambda$ is an intervention coefficient (set to $0.5$ by default).
In other words, we use SAE to dynamically modulate the strength of attention consistency enhancement for each head: heads with higher SAE (more uncertain intra-modal attention) are more strongly nudged towards the consensus pattern $C_k^{(l)}$, resulting in a more focused attention distribution over visual tokens.

Fig.~\ref{fig:sae_mitigation} presents a qualitative example of SAE-guided attention intervention.
Under greedy decoding without intervention, the model produces hallucinated objects that do not appear in the image.
After applying SAE-guided modulation, most hallucinated objects are removed or replaced, while real objects are better preserved or even completed. Correspondingly, the number of hallucinated objects decreases, whereas the number of real objects remains roughly unchanged, indicating that SAE-guided attention intervention effectively concentrates attention on image-consistent regions and suppresses object hallucinations.

\subsection{SAE-guided Embodied Navigation}
\label{sec:sae_guided_navigation}

LVLMs have been widely applied to embodied intelligence and robot navigation \cite{zu2024language,gu2022vision}. To show that SAE is useful not only for object hallucination detection but also for downstream multimodal decision-making, we apply the proposed inference-time intervention to an LVLM-based embodied navigation system. The target scenario involves a quadruped robot equipped with an RGB-D camera and LiDAR, where navigation decisions are made from first-person visual observations and high-level language instructions.

\begin{figure}[h]
    \centering
    \includegraphics[width=1.0\linewidth]{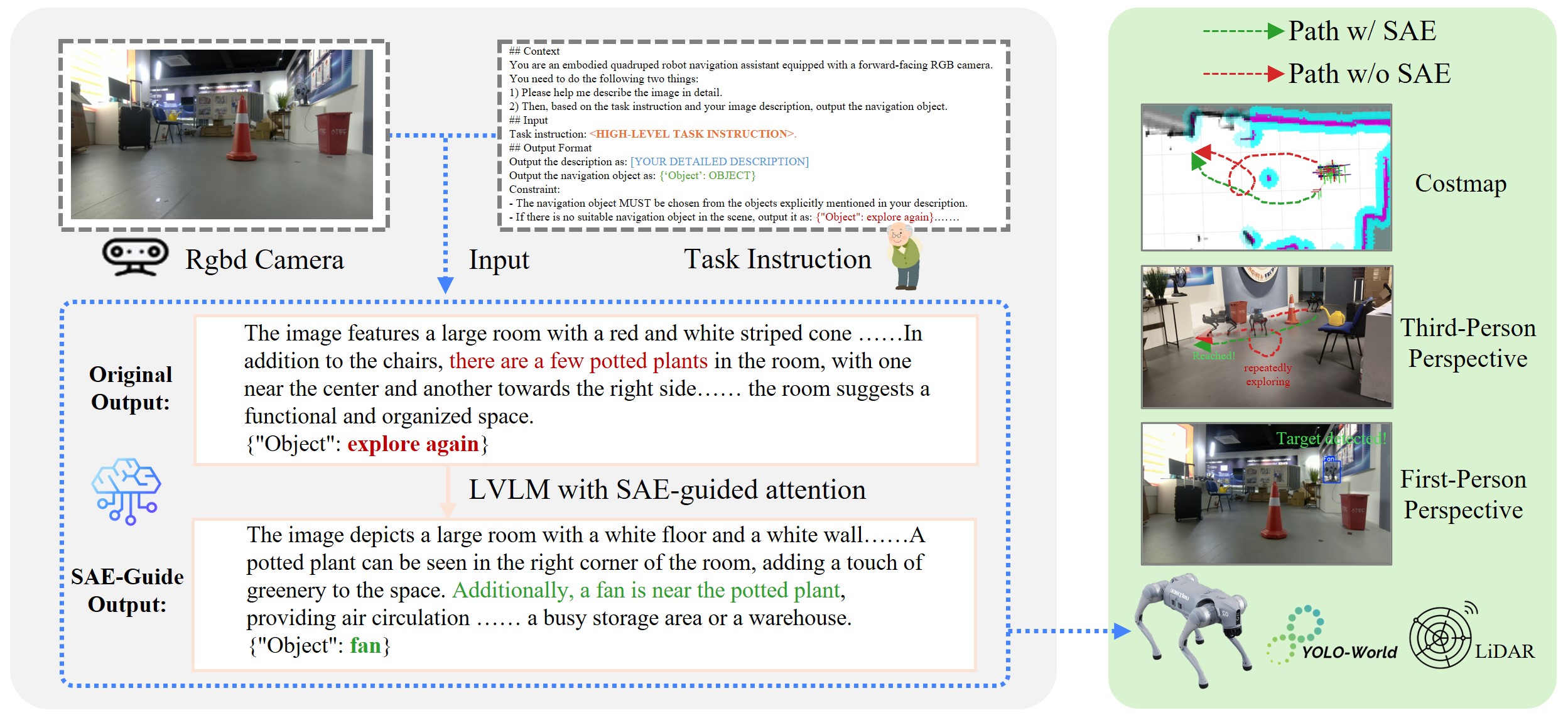}
    \caption{SAE-Guided Navigation Pipeline: The robot takes a first-person RGB-D image and language instruction as input to the LVLM, with the output being an action path in the physical world.}
    \label{fig:navigation}
\vspace{-3mm}
\end{figure}

Fig. \ref{fig:navigation} illustrates the overall pipeline. The robot receives a first-person image and a high-level instruction, while a navigation prompt guides the LVLM to describe the scene and output one or more target objects. Vanilla LVLMs are prone to hallucinations, often producing incorrect targets or vague ``explore again" instructions. To mitigate this issue, we adopt the method in Sec.~\ref{sec:sae_guided_mitigation} to reduce hallucinations. The predicted targets are then grounded in the image using YOLO-World \cite{cheng2024yolo}, an open-vocabulary detector. For each textual label, YOLO-World returns a bounding box, which is projected into the robot’s coordinate system using depth information and mapped onto a LiDAR-based local costmap as a waypoint. Finally, the local planner generates a collision-free path from the current pose to the target waypoint. This pipeline converts hallucination-prone LVLM predictions into a more reliable navigation process.

\section{Experiments}
\label{sec:experiments}

\subsection{Implementation Details}
\label{sec:implementation_details}

\textbf{Segmentation model.} We use Mask2Former-Swin-L~\cite{cheng2022masked} to produce semantic segmentation maps on the resized input image, and cache the masks for repeated evaluation.

\textbf{Token-to-segment assignment.} Let the image fed to the vision encoder be resized to a fixed resolution $R \times R$. For a ViT-style encoder with patch size $p$, each visual token $v_i$ corresponds to a patch receptive region $\Omega_i \subset [1,R]\times[1,R]$ of size $p \times p$. Given the segmentation masks $\{S_c\}_{c\in C}$, we assign each visual token the semantic class with maximum overlap inside its receptive region:
\begin{equation}
 c(i)=\arg\max_{c\in C}|\Omega_i \cap S_c|.
\label{eq:token_assignment}
\end{equation}
This majority-overlap rule yields a deterministic mapping from visual tokens to semantic categories.

\textbf{Attention mass and layer selection.}
For the generated token $y_k$, we define the total visual attention mass as
\begin{equation}
M_k^{(l,h)} = \sum_{i=1}^{n} A_k^{(l,h)}(a_k,i),
\label{eq:visual_mass}
\end{equation}
and use it in Eq.~\eqref{eq:reliability}. For reproducibility, we define the middle-layer set as the central 40\% of decoder blocks:
\begin{equation}
\mathcal{L}_{\mathrm{mid}}=
\left\{l \;\middle|\; \lceil 0.3L \rceil \le l \le \lfloor 0.7L \rfloor \right\}.
\label{eq:middle_layers}
\end{equation}
For LLaVA-1.5-7B ($L=32$), this corresponds to layers 11--22; for LLaVA-1.5-13B ($L=40$), it corresponds to layers 13--28.

\textbf{Multi-token object phrases.}
If a target phrase is split into multiple subword tokens, we compute its object-level score by averaging the scores over the corresponding subword positions.

\textbf{Default hyperparameters and devices.}
The intervention coefficient in Eq.~\eqref{eq:sae_update} is set to $\lambda=0.5$. All compared methods share the same prompt, temperature, and maximum generation length. All experiments were conducted on two NVIDIA 3090 (24GB) GPUs.

\begin{figure*}[!t]
\centering
\includegraphics[width=1.0\linewidth]{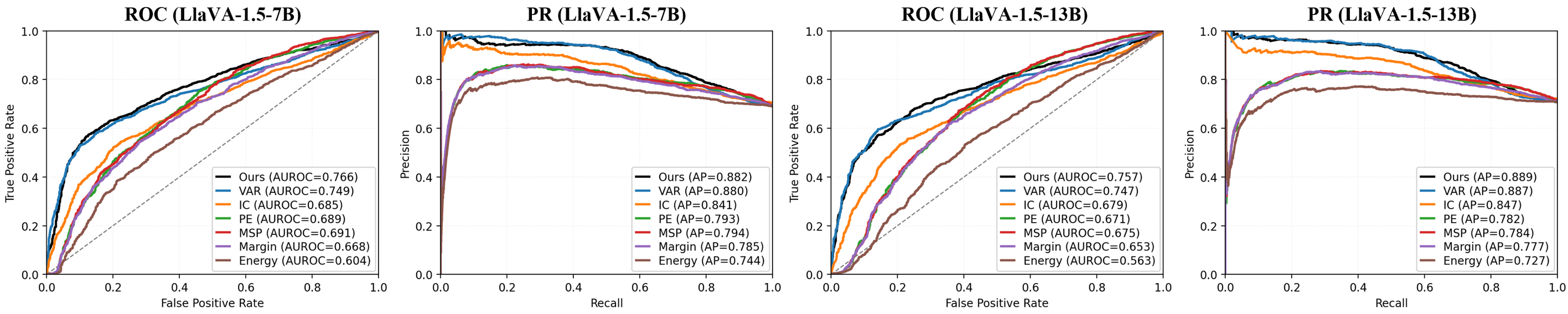}
\caption{ROC and PR curves of different hallucination detection methods on LLaVA-1.5-7B and LLaVA-1.5-13B.}
\label{fig:roc_pr}
\vspace{-3mm}
\end{figure*}

\subsection{Hallucination Detection}
\label{sec:hallucination_detection}

\textbf{Experimental setup.}
We evaluate two open-source LVLMs: LLaVA-1.5-7B and LLaVA-1.5-13B~\cite{liu2024improved}.
We randomly sample 1{,}000 images from the COCO 2014 validation set and let the models generate descriptions.
For each image, we extract object tokens using the CHAIR~\cite{rohrbach2018object} code and label them as real if the object matches the COCO ground truth of that image and hallucinated otherwise.
The basic detection unit is therefore the object token. For our approach, we use the SAE-based reliability score $R$ defined in Sec.~\ref{sec:SAE_for_Quantifying} as the detection score. For all methods, raw scores are converted when necessary so that a larger score consistently indicates a higher likelihood of real class.

\textbf{Evaluation metrics.}
We report AUROC and AUPRC \cite{Hanley1982AUROC,Davis2006PRvsROC}, standard metrics for binary detection. For a more intuitive comparison, we also plot full ROC and PR curves.

\textbf{Baselines.}
We compare uncertainty-based hallucination detection methods, including predictive entropy (PE) \cite{Li2024ReferenceFreeHallucination}, MSP \cite{Hendrycks2017BaselineMSP}, margin \cite{Tamang2024MarginBounded}, energy score \cite{Liu2020EnergyOOD}, internal confidence (IC) \cite{Jiang2025InternalConfidence}, and VAR \cite{Jiang2025DevilsMiddleLayers}, using their original implementations.

Fig.~\ref{fig:roc_pr} shows that our method achieves competitive detection performance on both models and generally outperforms the compared baselines in the ROC and PR curves, with the advantage being more noticeable in the low-false-positive and high-recall regions.
In terms of summary metrics, on LLaVA-1.5-7B our method achieves an AUROC/AUPRC of 0.766/0.882, compared with 0.749/0.880 for the strongest baseline VAR; on LLaVA-1.5-13B, the corresponding scores improve from 0.747/0.887 to 0.757/0.889.
Compared with output-distribution-based uncertainty measures (PE, MSP, Margin, and Energy), our method yields clear gains in AUROC and also improves AUPRC.
Overall, these results suggest that modeling intra-visual uncertainty via SAE provides a useful and complementary signal for distinguishing hallucinated from real objects.

\subsection{Hallucination Mitigation}
\label{sec:hallucination_mitigation}

\textbf{Experimental setup.}
We reuse the two LVLMs and the COCO 2014 setup from Sec.~\ref{sec:hallucination_detection}.
Decoding temperature, maximum generation length, and other hyperparameters are identical across methods, with all methods using the same prompts.
Our method applies the SAE-guided inference-time attention intervention in Sec.~\ref{sec:sae_guided_mitigation}.

\textbf{Evaluation metrics.}
We use CHAIR~\cite{rohrbach2018object}, a standard hallucination benchmark, reporting sentence-level ($C_S$) and instance-level ($C_I$) rates; lower is better. Following prior LVLM studies, we compute precision, recall, and F1 with ground-truth objects as positives.

\textbf{Baselines.}
We consider three decoding strategies: greedy decoding, beam search~\cite{freitag2017beam}, and nucleus (top-$p$) sampling~\cite{Holtzman2020NeuralTextDegeneration}. We also evaluate four inference-time hallucination mitigation methods: OPERA \cite{Huang2024OPERA} augments beam search with an over-trust penalty and retrospection-allocation; VCD \cite{Leng2024VCD} performs visual contrastive decoding; PAI \cite{Liu2025PAI} strengthens visual attention in a training-free manner; and VAR \cite{Jiang2025DevilsMiddleLayers} rescales attention using middle-layer statistics. All intervention-based methods are built on greedy decoding.

\begin{table}[t]
\centering
\caption{CHAIR and P/R/F1 results of different methods on LLaVA-1.5-7B and LLaVA-1.5-13B (max length = 512).}
\vspace{-3mm}
\label{tab:chair_main}
\setlength{\tabcolsep}{3pt}
\renewcommand{\arraystretch}{1.15}
\resizebox{\linewidth}{!}{
\begin{tabular}{lcccccccccccc}
\toprule
\multirow{2}{*}{Method} &
\multicolumn{5}{c}{LLaVA-1.5-7B} &
\multicolumn{5}{c}{LLaVA-1.5-13B} &
\multicolumn{2}{c}{Avg} \\
\cmidrule(lr){2-6} \cmidrule(lr){7-11} \cmidrule(lr){12-13}
& $C_S\!\downarrow$ & $C_I\!\downarrow$ & R$\uparrow$ & P$\uparrow$ & F1$\uparrow$
& $C_S\!\downarrow$ & $C_I\!\downarrow$ & R$\uparrow$ & P$\uparrow$ & F1$\uparrow$
& $C_S\!\downarrow$ & $C_I\!\downarrow$ \\
\midrule
Nucleus & 56.0 & 16.4 & 73.4 & 71.0 & 72.2 & 54.0 & 15.0 & 74.6 & 73.0 & 73.8 & 55.0 & 15.7 \\
Greedy  & 47.9 & 13.6 & 78.1 & 75.8 & 77.0 & 44.3 & 11.7 & 77.6 & 78.7 & \textbf{78.2} & 46.1 & 12.6 \\
Beam    & 52.1 & 14.1 & \textbf{78.5} & 75.2 & 76.8 & 49.7 & 12.7 & \textbf{79.1} & 77.3 & \textbf{78.2} & 50.9 & 13.4 \\
OPERA   & 48.5 & 13.7 & 78.1 & 75.5 & 76.8 & 52.6 & 14.1 & 68.1 & 73.1 & 70.5 & 50.6 & 13.9 \\
\hdashline[1pt/2pt]
VCD     & 55.4 & 15.7 & 75.6 & 71.8 & 73.7 & 52.3 & 13.9 & 75.5 & 74.6 & 75.1 & 53.8 & 14.8 \\
PAI     & 30.1 &  8.8 & 70.8 & 83.7 & 76.7 & 28.9 & \textbf{8.5} & 70.4 & 83.4 & 76.4 & 29.5 &  8.6 \\
VAR     & 25.3 &  6.5 & 69.2 & 86.6 & \textbf{76.9} & 26.7 &  8.8 & 67.3 & 84.1 & 74.7 & 26.0 &  7.6 \\
\rowcolor{gray!20}
Ours    & \textbf{24.2} & \textbf{5.9} & 68.2 & \textbf{87.3} & 76.6
        & \textbf{23.8} & \textbf{8.5} & 64.8 & \textbf{85.9} & 73.9
        & \textbf{24.0} & \textbf{7.2} \\
\bottomrule
\end{tabular}
} 
\vspace{-3mm}
\end{table}

Tab.~\ref{tab:chair_main} shows the upper block as decoding strategies and the lower as inference-time interventions. Our method achieves the lowest or tied‑for‑lowest sentence- and instance-level hallucination rates $C_S$ and $C_I$ on both models with F1 close to the best baselines. On LLaVA-1.5-7B, greedy decoding yields $C_S/C_I$ of 47.9/13.6, whereas our method reduces them to 24.2/5.9, roughly halving hallucinations; on LLaVA-1.5-13B, $C_S$ similarly drops from 44.3 to 23.8. Overall, SAE-guided attention intervention provides the best average $C_S$ and $C_I$ with only a small F1 loss, achieving a favorable trade-off between low hallucination and high accuracy.

Tab.~\ref{tab:chair_length} further studies different maximum generation lengths (64/128/256 tokens) on LLaVA-1.5-7B.
For most methods, F1 increases from 64 to 128 tokens and then remains relatively stable, while $C_S$ and $C_I$ consistently rise with larger generation budgets.
Our method consistently achieves the lowest $C_S$ and $C_I$ across all lengths (24.2/6.4 at 256 tokens), with slower growth than the compared baselines, while maintaining competitive F1 (75.5 on average).
These results suggest that SAE-guided attention intervention remains effective under different generation budgets and is relatively robust to output length.

\begin{table}[t]
\centering
\caption{Effect of maximum generation length on hallucination (CHAIR) and F1 on LLaVA-1.5-7B.}
\vspace{-3mm}
\label{tab:chair_length}
\setlength{\tabcolsep}{3pt}
\renewcommand{\arraystretch}{1.15}
\resizebox{\linewidth}{!}{%
\begin{tabular}{lcccccccccccc}
\toprule
\multirow{2}{*}{Method} &
\multicolumn{3}{c}{64} &
\multicolumn{3}{c}{128} &
\multicolumn{3}{c}{256} &
\multicolumn{3}{c}{Avg} \\
\cmidrule(lr){2-4} \cmidrule(lr){5-7} \cmidrule(lr){8-10} \cmidrule(lr){11-13}
& $C_S\!\downarrow$ & $C_I\!\downarrow$ & F1$\uparrow$
& $C_S\!\downarrow$ & $C_I\!\downarrow$ & F1$\uparrow$
& $C_S\!\downarrow$ & $C_I\!\downarrow$ & F1$\uparrow$
& $C_S\!\downarrow$ & $C_I\!\downarrow$ & F1$\uparrow$ \\
\midrule
Nucleus & 25.6 &  9.2 & 68.7 & 54.1 & 16.1 & 72.0 & 56.0 & 16.4 & 72.2 & 45.2 & 13.9 & 71.0 \\
Greedy  & 21.3 &  6.6 & \textbf{74.5} & 46.9 & 13.3 & \textbf{77.0} & 47.9 & 13.6 & \textbf{77.0} & 38.7 & 11.2 & \textbf{76.2} \\
Beam    & 18.7 &  6.1 & 73.3 & 50.2 & 13.7 & 76.9 & 52.1 & 14.1 & 76.8 & 40.3 & 11.3 & 75.7 \\
OPERA   & 17.0 &  5.4 & 74.3 & 47.2 & 13.4 & 76.8 & 48.5 & 13.7 & 76.8 & 37.6 & 10.8 & 76.0 \\
\hdashline[1pt/2pt]
VCD     & 23.2 &  8.1 & 71.9 & 53.9 & 15.0 & 74.6 & 55.4 & 15.7 & 73.7 & 44.2 & 12.9 & 73.4 \\
PAI     & 19.2 &  6.1 & 73.7 & 30.0 &  8.8 & 76.7 & 30.1 &  8.8 & 76.7 & 26.4 &  7.9 & 75.7 \\
VAR     & 16.6 &  5.0 & 73.5 & 25.2 &  6.7 & 76.9 & 25.3 &  6.7 & 76.9 & 22.4 &  6.1 & 75.8 \\
\rowcolor{gray!20}
Ours    & \textbf{16.4} & \textbf{4.9} & 73.7
        & \textbf{21.7} & \textbf{6.0} & 76.2
        & \textbf{24.2} & \textbf{6.4} & 76.6
        & \textbf{20.8} & \textbf{5.8} & 75.5 \\
\bottomrule
\end{tabular}
}
\vspace{-3mm}

\end{table}

\subsection{Cross-family \& Dataset Generalization}
\label{sec:cross_family_cross_dataset}

\textbf{Additional LVLM families.}
We test Qwen2-VL-7B and InternVL2-8B~\cite{wang2024qwen2vl,chen2024internvl2}, which differ substantially from LLaVA-1.5 in visual tokenization, resolution handling, and multimodal fusion design.

\textbf{Additional benchmarks.}
Besides COCO-CHAIR, we evaluate on two widely used hallucination benchmarks. (1) \textbf{POPE}~\cite{li2023pope}, which probes object existence under random, popular, and adversarial negative sampling; we report average Accuracy and F1 over the three official splits. (2) \textbf{MMHal-Bench}~\cite{sun2024mmhal}, which evaluates open-ended multimodal hallucinations over diverse question types; we report GPT-based score and hallucination rate. Together with COCO-CHAIR, these benchmarks cover caption-based evaluation, yes/no probing, and open-ended factuality.

\textbf{Evaluation protocol.}
For POPE, we follow the official protocol and answer each yes/no question with greedy decoding. For MMHal-Bench, SAE-guided intervention is applied during the full answer generation process. We compare against Greedy, VCD, PAI, and VAR under matched prompts and decoding settings.

\begin{table}[t]
\centering
\caption{\textbf{POPE and MMHal-Bench results on LLaVA-1.5-7B.}}
\vspace{-3mm}
\label{tab:cross_family_cross_dataset}
\setlength{\tabcolsep}{4.8pt}
\renewcommand{\arraystretch}{1.12}
\resizebox{\linewidth}{!}{
\begin{tabular}{lcccc}
\toprule
Method & POPE Acc. $\uparrow$ & POPE F1 $\uparrow$ & MMHal Score $\uparrow$ & MMHal Hall. Rate $\downarrow$ \\
\midrule
Greedy & 84.7 & 85.2 & 2.58 & 55.4  \\
VCD    & 86.0 & 86.4 & 2.71 & 52.1  \\
PAI    & 87.1 & 87.5 & 2.87 & 49.1 \\
VAR    & 87.6 & 88.0 & 2.95 & 47.4  \\
\rowcolor{gray!20}
Ours   & \textbf{88.8} & \textbf{89.1} & \textbf{3.11} & \textbf{43.2} \\
\bottomrule
\end{tabular}}
\vspace{-3mm}

\end{table}

\begin{table}[t]
\centering
\footnotesize
\caption{CHAIR results on Qwen2-VL-7B and InternVL2-8B.}
\vspace{-3mm}
\label{tab:cross_family_coco}
\setlength{\tabcolsep}{3pt}
\renewcommand{\arraystretch}{1.15}
{
\begin{tabular}{lcccccc}
\toprule
\multirow{2}{*}{Method} &
\multicolumn{3}{c}{Qwen2-VL-7B} &
\multicolumn{3}{c}{InternVL2-8B} \\
\cmidrule(lr){2-4} \cmidrule(lr){5-7}
& $C_S\!\downarrow$ & $C_I\!\downarrow$ & F1$\uparrow$
& $C_S\!\downarrow$ & $C_I\!\downarrow$ & F1$\uparrow$ \\
\midrule
Nucleus & 36.8 & 10.4 & 72.9 & 38.4 & 10.9 & 73.4 \\
Greedy  & 31.6 &  9.1 & \textbf{77.4} & 33.8 &  9.6 & \textbf{77.1} \\
Beam    & 33.0 &  9.3 & 77.1 & 35.1 &  9.8 & 76.8 \\
OPERA   & 32.8 &  9.1 & 76.6 & 34.6 &  9.7 & 76.2 \\
\hdashline[1pt/2pt]
VCD     & 35.9 & 10.1 & 74.6 & 37.2 & 10.4 & 74.2 \\
PAI     & 26.2 &  7.3 & 76.9 & 27.6 &  7.8 & 76.5 \\
VAR     & 24.4 &  6.8 & 76.8 & 25.7 &  7.1 & 76.3 \\
\rowcolor{gray!20}
Ours    & \textbf{21.9} & \textbf{6.0} & 76.7
        & \textbf{23.1} & \textbf{6.4} & 76.2 \\
\bottomrule
\end{tabular}}
\vspace{-3mm}

\end{table}

Tab.~\ref{tab:cross_family_cross_dataset} and Tab.~\ref{tab:cross_family_coco} suggest that SAE transfers beyond the original COCO/LLaVA setting. On POPE and MMHal-Bench with LLaVA-1.5-7B, our method improves both probing-based and open-ended hallucination metrics. On Qwen2-VL-7B and InternVL2-8B, it consistently reduces CHAIR scores while remaining competitive in F1. These results provide initial evidence that SAE is not limited to a single benchmark or a single LVLM family.

\subsection{LVLM-based Navigation}
\label{sec:LVLM-based_Navigation}

\textbf{Scene setup.}
We conduct a real-robot study in indoor environments using a Unitree Go1 quadruped robot equipped with an RGB-D camera and LiDAR. LLaVA-1.5-7B serves as the high-level perception and decision-making module, and YOLO-World~\cite{cheng2024yolo} is used to ground open-vocabulary targets in the image. Grounded detections are projected into the robot coordinate system using depth information, and LiDAR scans are used to build a local costmap for path planning.

\begin{figure*}[t]
\centering
\includegraphics[width=0.90\linewidth]{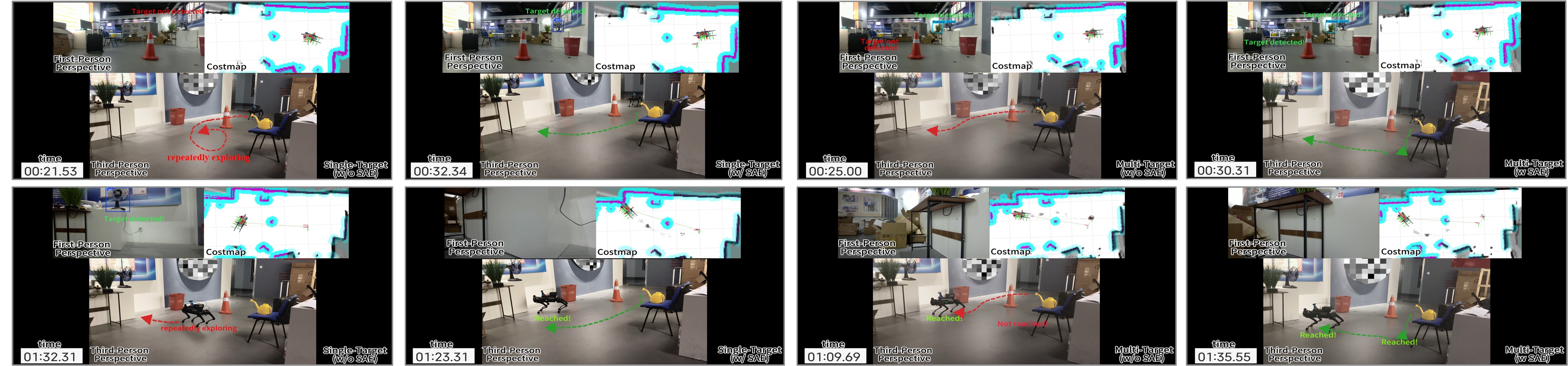}
\caption{LVLM navigation in single- and multi-target scenarios with and without SAE.
}
\label{fig:experimental_navigation}
\end{figure*}

\textbf{Quantitative protocol.}
Beyond the qualitative examples, we conduct a quantitative study in two settings: single-target navigation, where the robot is instructed to reach one object, and multi-target navigation, where it must visit two target objects sequentially. For each setting, we define 12 episode templates across four indoor scenes with randomized initial poses and language instructions, and execute each template for 10 independent runs. We compare the vanilla pipeline with SAE-guided intervention under the same protocol. We report Success Rate, Episode Time (s), Replans/Episode, and Unsupported Targets/Episode, where the last metric denotes the average number of LVLM target proposals that cannot be grounded in the current scene. In the multi-target setting, episode time should be interpreted together with the success rate rather than in isolation.

\begin{table}[t]
\centering
\footnotesize
\caption{Real-robot study on LVLM-based navigation.}
\vspace{-3mm}
\label{tab:robot_quant}
\setlength{\tabcolsep}{2.2pt}
\renewcommand{\arraystretch}{1.05}
\begin{tabular}{@{}lccccc@{}}
\toprule
Scenario & Method & Success Rate$\uparrow$ & Time (s) & Replans$\downarrow$ & Unsup.$\downarrow$ \\
\midrule
\multirow{2}{*}{Single-target}
& Vanilla & 66.7 & 96.4 $\pm$ 18.7 & 2.3 $\pm$ 1.0 & 0.92 $\pm$ 0.51 \\
& Ours    & \textbf{91.7} & \textbf{84.1 $\pm$ 16.2} & \textbf{1.1 $\pm$ 0.6} & \textbf{0.25 $\pm$ 0.22} \\
\midrule
\multirow{2}{*}{Multi-target}
& Vanilla & 41.7 & 72.6 $\pm$ 13.9 & 4.0 $\pm$ 1.4 & 1.58 $\pm$ 0.67 \\
& Ours    & \textbf{75.0} & 99.8 $\pm$ 15.6 & \textbf{2.1 $\pm$ 1.0} & \textbf{0.63 $\pm$ 0.41} \\
\bottomrule
\end{tabular}
\vspace{-3mm}

\end{table}

Tab.~\ref{tab:robot_quant} shows that SAE-guided intervention consistently improves success rate and reduces replans and unsupported target proposals in both settings. In the single-target setting, it also shortens the average episode time by reducing repeated exploration caused by unsupported targets. In the multi-target setting, the average episode time becomes longer, because the SAE-guided agent more often continues to the correct second target and completes the intended sequence, whereas the vanilla pipeline frequently terminates early after missing an intermediate target. We therefore interpret the multi-target time together with the success rate rather than as a standalone efficiency metric. We view these results as preliminary evidence that SAE can improve the reliability of LVLM-driven navigation. These quantitative observations are consistent with the qualitative examples in Fig.~\ref{fig:experimental_navigation}.

\subsection{Ablation Studies}
\label{sec:additional_ablation}

We conduct ablation studies to examine the sensitivity, robustness, region-partition dependence, and computational cost of SAE.

\textbf{Sensitivity to the intervention strength.}
Tab.~\ref{tab:lambda_ablation} studies the intervention coefficient $\lambda$. Increasing $\lambda$ initially improves hallucination suppression, but overly large values sharply reduce recall and F1, indicating under-generation rather than better grounding. In particular, excessively strong intervention can artificially lower CHAIR by suppressing object generation itself. We therefore set $\lambda=0.5$ by default, which provides a good trade-off between hallucination reduction and caption completeness.

\begin{table}[t]
\centering
\scriptsize
\setlength{\tabcolsep}{2.8pt}
\renewcommand{\arraystretch}{0.9}
\caption{Sensitivity to $\lambda$ on LLaVA-1.5-7B (CHAIR, P/R/F1).}
\vspace{-3mm}
\label{tab:lambda_ablation}
\begin{tabular}{@{}c|ccccc@{}}
\toprule
$\lambda$ (Ours) & $C_S$ $\downarrow$ & $C_I$ $\downarrow$ & R $\uparrow$ & P $\uparrow$ & F1 $\uparrow$ \\
\midrule
0.3 & 40.4 & 11.8 & 75.5 & 77.3 & 76.4 \\
0.4 & 40.2 & 12.1 & 75.1 & 76.1 & 75.6 \\
0.5 & 24.2 & 5.9  & 68.2 & 87.3 & 76.6 \\
0.6 & 2.6  & 2.0  & 37.8 & 97.8 & 54.6 \\
0.7 & 1.4  & 1.8  & 8.9  & 98.0 & 16.2 \\
\midrule
Greedy & 47.9 & 13.6 & 78.1 & 75.8 & 77.0 \\
\bottomrule
\end{tabular}
\vspace{-3mm}
\end{table}

\textbf{Robustness to segmentation errors.}
SAE operates on token-level region assignments induced by semantic segmentation, rather than on pixel-accurate masks. We therefore evaluate its robustness under three controlled perturbation settings: boundary perturbation of the segmentation masks, random token-to-segment reassignment, and coarsened region partitions.

\begin{table}[t]
\centering
\scriptsize
\setlength{\tabcolsep}{3.2pt}
\renewcommand{\arraystretch}{0.95}
\caption{Robustness to segmentation errors on LLaVA-1.5-7B.}
\vspace{-3mm}

\label{tab:seg_noise_ablation}
\begin{tabular}{@{}l|ccccc@{}}
\toprule
Partition setting & $C_S$ $\downarrow$ & $C_I$ $\downarrow$ & R $\uparrow$ & P $\uparrow$ & F1 $\uparrow$ \\
\midrule
Greedy & 47.9 & 13.6 & 78.1 & 75.8 & 77.0 \\
Ours (clean seg.) & 24.2 & 5.9 & 68.2 & 87.3 & 76.6 \\
\midrule
Boundary noise (mild) & 25.1 & 6.2 & 67.9 & 86.8 & 76.1 \\
Boundary noise (moderate) & 26.4 & 6.7 & 67.1 & 86.1 & 75.4 \\
Token reassignment ($\rho=0.1$) & 25.7 & 6.3 & 67.6 & 86.5 & 75.9 \\
Token reassignment ($\rho=0.2$) & 27.3 & 7.1 & 66.7 & 85.4 & 74.9 \\
Coarse segmentation ($\times 4$) & 26.1 & 6.5 & 67.4 & 86.0 & 75.6 \\
\bottomrule
\end{tabular}
\vspace{-3mm}

\end{table}

Tab.~\ref{tab:seg_noise_ablation} shows that moderate partition errors lead to only modest degradation across both hallucination and accuracy metrics. These results indicate that SAE depends primarily on a stable token-level region partition rather than on pixel-accurate boundaries.

\textbf{Ablation on region partitioning methods.}
To disentangle the effect of semantic grouping from that of generic region-level aggregation, we compare the default semantic partition based on Mask2Former-Swin-L with a lighter Mask2Former-Swin-T variant and a class-agnostic block-based partition. Block entropy is a lightweight method that computes entropy over spatially contiguous visual-token blocks without semantic labels, capturing region-level aggregation but unable to merge spatially disjoint regions from the same semantic category.

\begin{table}[t]
\centering
\scriptsize
\setlength{\tabcolsep}{3.0pt}
\renewcommand{\arraystretch}{0.95}
\caption{Ablation on region partitioning on LLaVA-1.5-7B.}
\vspace{-3mm}

\label{tab:seg_backbone_ablation}
\begin{tabular}{@{}l|cccc@{}}
\toprule
Setting & $C_S$ $\downarrow$ & $C_I$ $\downarrow$ & F1 $\uparrow$ & Time / image (s) $\downarrow$ \\
\midrule
Greedy (no partition) & 47.9 & 13.6 & 77.0 & 0.82 \\
Block entropy & 37.4 & 11.2 & 76.2 & 0.93 \\
Mask2Former-Swin-T & 27.8 & 7.2 & 75.0 & 1.01 \\
Mask2Former-Swin-L & \textbf{24.2} & \textbf{5.9} & \textbf{76.6} & 1.17 \\
\bottomrule
\end{tabular}
\vspace{-3mm}

\end{table}

Tab.~\ref{tab:seg_backbone_ablation} shows that our SAE design with semantic labels delivers the strongest mitigation performance, while a cheaper class-agnostic block-based approximation still provides partial gains over greedy decoding through region-level grouping, further highlighting the effectiveness and practical value of our semantic region-aware aggregation. Replacing Mask2Former-Swin-L with the smaller Mask2Former-Swin-T slightly worsens CHAIR and F1, but still outperforms region-connectivity-based partitioning and greedy decoding. Overall, semantic partitioning is more effective than using region structure alone, yielding additional gains by more reliably grouping semantically coherent yet spatially separated regions.

\textbf{Runtime overhead.}
We report runtime overhead under several settings on LLaVA-1.5-7B, using end-to-end token generation speed as the main efficiency metric and peak GPU memory as a complementary cost indicator. All settings use the same input resolution and greedy decoding configuration.

\begin{table}[t]
\centering
\scriptsize
\setlength{\tabcolsep}{3.0pt}
\renewcommand{\arraystretch}{0.95}
\caption{\textbf{Runtime overhead analysis} on LLaVA-1.5-7B.}
\vspace{-3mm}

\label{tab:runtime_ablation}
\begin{tabular}{@{}l|ccc@{}}
\toprule
Method & Tokens/sec.$\uparrow$ & Relative overhead$\downarrow$ & GPU memory (GB) \\
\midrule
Greedy & 17.3 & 1.00$\times$ & 21.4 \\
Ours (precomputed masks) & 16.5 & 1.05$\times$ & 21.8 \\
Ours + block entropy & 16.2 & 1.07$\times$ & 21.9 \\
Ours + online Swin-T & 15.4 & 1.12$\times$ & 22.4 \\
Ours + online Swin-L & 14.4 & 1.20$\times$ & 23.1 \\
\bottomrule
\end{tabular}
\vspace{-3mm}

\end{table}

Tab.~\ref{tab:runtime_ablation} shows that SAE maintains competitive efficiency across different partition settings. With precomputed masks, our method remains close to greedy decoding, and the block-entropy variant preserves a similarly lightweight runtime profile while retaining strong mitigation performance. Online semantic partitioning introduces additional cost, but the smaller Swin-T extractor offers a favorable efficiency--performance trade-off, showing that SAE can be deployed effectively under different latency budgets.

\textbf{Ablation on selected layers.}
We also compare early-, middle-, late-, and all-layer aggregation strategies. Tab.~\ref{tab:layer_ablation} shows that middle-layer aggregation yields the best overall trade-off, achieving the lowest hallucination rates and the highest precision/F1, while all-layer aggregation slightly improves recall.

\begin{table}[t]
\centering
\scriptsize
\setlength{\tabcolsep}{3.0pt}
\renewcommand{\arraystretch}{0.95}
\caption{Ablation on layer selection on LLaVA-1.5-7B.}
\vspace{-3mm}

\label{tab:layer_ablation}
\begin{tabular}{@{}l|ccccc@{}}
\toprule
Layer range & $C_S$ $\downarrow$ & $C_I$ $\downarrow$ & R $\uparrow$ & P $\uparrow$ & F1 $\uparrow$ \\
\midrule
Early layers (1--10) & 31.8 & 8.7 & 70.5 & 82.1 & 75.9 \\
Middle layers (11--22) & \textbf{24.2} & \textbf{5.9} & 68.2 & \textbf{87.3} & \textbf{76.6} \\
Late layers (23--32) & 29.7 & 7.6 & 66.1 & 84.8 & 74.3 \\
All layers & 27.9 & 7.1 & \textbf{71.0} & 84.1 & 76.1 \\
\bottomrule
\end{tabular}
\vspace{-5mm}
\end{table}

Taken together, the above ablations suggest that SAE is not tied to pixel-accurate masks or to a single high-capacity segmentor. Its main requirement is a stable region partition over visual tokens. The class-agnostic block-entropy variant already preserves much of the mitigation benefit, suggesting that region-level attention concentration is a key factor. At the same time, stronger semantic partitions remain useful for grouping semantically coherent regions, which is consistent with the remaining gap between block entropy and full SAE. From an efficiency perspective, most overhead comes from the partition extractor rather than from the intervention itself, and lighter partition strategies offer attractive trade-offs between computational cost and mitigation quality.

\section{Conclusion}

We proposed Segmentation-based Attention Entropy, a semantic uncertainty measure for characterizing how diffusely an LVLM allocates visual attention across segmentation-derived object categories during generation. Our results show that SAE serves as an effective signal for both hallucinated-object detection and training-free inference-time mitigation. Beyond COCO captioning, we validated the effectiveness and generality of SAE across multiple LVLM families, diverse hallucination-oriented benchmarks, and real-world embodied navigation scenarios. Overall, these findings suggest that semantic concentration in visual attention is a useful and interpretable cue for improving LVLM factuality and reliability. In future work, we will extend SAE to more challenging multi-object scenes, phrase-level grounding tasks, and embodied reasoning settings.

\begin{acks}
This work was supported by the
\grantsponsor{NSFC}
{National Natural Science Foundation of China}{}
(No.~\grantnum{NSFC}{62176052}),
the
\grantsponsor{FRFCU}
{Fundamental Research Funds for the Central Universities}{}
(\grantnum{FRFCU}{DHU Distinguished Young Professor Program}),
and the
\grantsponsor{SMEC}
{AI-Enhanced Research Program of the Shanghai Municipal Education Commission}{}
(\grantnum{SMEC}{SMEC-AI-DHUZ-05}).
\end{acks}

\bibliographystyle{ACM-Reference-Format}
\bibliography{ref}

\end{document}